\documentclass[conference]{IEEEtran}
\IEEEoverridecommandlockouts
\usepackage{cite}
\usepackage{amsmath,amssymb,amsfonts}
\usepackage{algorithmic}
\usepackage{graphicx}
\usepackage{textcomp}
\usepackage{xcolor}
\usepackage[hidelinks,bookmarks=false]{hyperref}
\usepackage{tikz}
\usepackage[inline]{enumitem}
\usepackage{pifont} 
\usepackage{listings}
\graphicspath{ {./images/} }
\def\BibTeX{{\rm B\kern-.05em{\sc i\kern-.025em b}\kern-.08em
    T\kern-.1667em\lower.7ex\hbox{E}\kern-.125emX}}
\begin{document}
\bstctlcite{IEEEexample:BSTcontrol}

\newcommand*\circled[1]{\tikz[baseline=(char.base)]{
        \node[shape=circle,draw,inner sep=.6pt] (char) {#1};}}
\newcommand{\FedCampus}{\textsc{FedCampus}}
\newcommand{\FedKit}{\textsc{FedKit}}
\newcommand{\challa}{\textbf{(C0)}}
\newcommand{\challb}{\textbf{(C1)}}
\newcommand{\challc}{\textbf{(C2)}}

\setlength\floatsep{0.5\baselineskip plus 1pt minus 2pt}
\setlength\textfloatsep{0.5\baselineskip plus 1pt minus 2pt}
\setlength\intextsep{0.5\baselineskip plus 1pt minus 2 pt}
\renewcommand\topfraction{0.95}
\renewcommand\bottomfraction{0.95}
\renewcommand\textfraction{0.05}
\renewcommand\floatpagefraction{0.95}
\setlength{\abovecaptionskip}{3pt plus 0pt minus 2pt}
\setlength{\skip\footins}{0.3cm}

\abovedisplayshortskip=3pt
\belowdisplayshortskip=3pt
\abovedisplayskip=5pt
\belowdisplayskip=5pt

\title
{\FedKit{}: Enabling Cross-Platform Federated Learning for Android and iOS}

\author{
\IEEEauthorblockN{Sichang He\IEEEauthorrefmark{1}, Beilong Tang\IEEEauthorrefmark{1}, Boyan Zhang\IEEEauthorrefmark{1}, Jiaoqi Shao\IEEEauthorrefmark{1}\IEEEauthorrefmark{2}, Xiaomin Ouyang\IEEEauthorrefmark{3}, Daniel\,Nata Nugraha\IEEEauthorrefmark{4}, Bing Luo\IEEEauthorrefmark{1}}
    \IEEEauthorblockA{\IEEEauthorrefmark{1}
        Duke Kunshan University, Jiangsu, China,
        \IEEEauthorrefmark{2}
        The Hong Kong University of Science and Technology, China,\\
        \IEEEauthorrefmark{3}
        University of California, Los Angeles, USA,
        \IEEEauthorrefmark{4}Flower Labs GmbH, Winterhuder Weg 29, 22085 Hamburg, Germany
    }
\thanks{This work of Sichang He, Beilong Tang, and Boyan Zhang was supported by DKU Undergraduate Studies Office through the Summer Research Scholars program. The work of Jiaqi Shao and Bing Luo was supported by the Suzhou Frontier Science and Technology Program (SYG202310). (Corrsponding author: Bing Luo.)}}

\maketitle
\begin{abstract}
We present \FedKit{}, a federated learning (FL) system tailored for
cross-platform FL research on \textit{Android and iOS} devices.
\FedKit{} pipelines cross-platform FL development by
enabling model conversion,
hardware-accelerated training,
and cross-platform model aggregation.
Our FL workflow supports flexible machine learning operations (MLOps) in production,
facilitating continuous model delivery and training.
We have deployed \FedKit{} in a real-world use case for
health data analysis on university campuses,
demonstrating its effectiveness.
\FedKit{} is open-source at \url{https://github.com/FedCampus/FedKit}.
\end{abstract}


\section{Motivation and Analysis}
FL is promising for training shared ML models collaboratively on
end devices while preserving data privacy~\cite{farcas2023demo}.
Yet, most FL research relies on simulations on desktop computers,
which may overlook constraints in realistic FL applications.
To enhance FL algorithm design with real-world data,
we aim to build a practical mobile FL system
harnessing real user data.

However,
existing accessible mobile FL systems exhibit important limitations,
as outlined in Table~\ref{tbl:fn-systems}.
Specifically, we identify \textbf{three key challenges}:
\challa~To collaboratively train the same models across
our users' diverse smartphones,
we require \textbf{cross-platform on-device training and model aggregation}.
\challb~To customize FL algorithms and update models in production,
we need \textbf{maximal control over the FL process} on
user' devices from our end.
\challc~The unfamiliar mobile \textbf{development environment} hinders
data scientists while developing models for mobile FL.

\section{Proposed Solution}

\FedKit{} is designed to enable practical \textbf{cross-platform} FL research on
\textbf{Android and iOS}.
In Sec.~\ref{sec:pipeline},
we present an FL pipeline that converts Python-based models,
and trains and aggregates them across platforms,
addressing~\challa{}.
For~\challb{},
our FL workflow enables \textbf{flexible MLOps} from the backend in production,
as we detail in Sec.~\ref{sec:mlops}.
The entire procedure tackles~\challc{} by
ensuring a user-friendly development environment in Python.
Overall,
\FedKit{} facilitates FL across Android and iOS client devices,
coordinated by a single Backend server.
Each client trains a local model with private data,
and the Backend performs cross-platform aggregation of these local models to
update the global model.

\subsection{Cross-Platform FL Model Pipeline}
\label{sec:pipeline}

\begin{figure}
    \centering
    \includegraphics*[width=\linewidth]{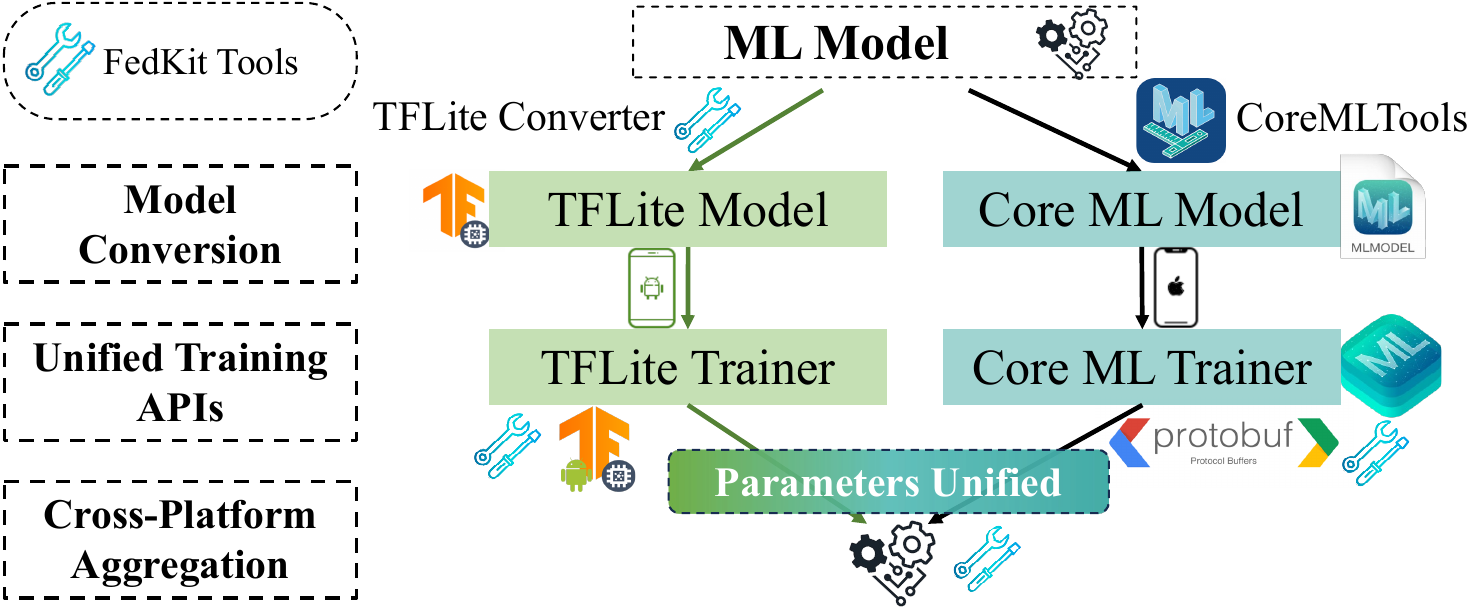}
    \caption{\FedKit{} Model Pipeline for Cross-Platform FL.}
    \label{cross_fl}
        \vspace{-1mm}
\end{figure}

To enable cross-platform FL,
especially \textit{cross-platform aggregation},
we propose a pipeline comprising
\textit{model conversion} and
\textit{unified training APIs},
as shown in Fig.~\ref{cross_fl}.

\begin{table}
    \centering
    \newcommand{\Ys}{\ding{51}}
    \newcommand{\No}{\ding{55}}
\begin{tabular}{lcccccc}
Functionality               & \cite{he2020fedml}
                                    & \cite{madrigal2023project}
                                            & \cite{mathur2021ondevice}
                                                    & \cite{hall2021syft}
                                                            & \FedKit{}\\
\hline
Android-Only                & \Ys   & \Ys   & \Ys   & \Ys   & \Ys   \\
iOS-Only                    & \No   & \No   & \Ys   & \Ys   & \Ys   \\
Cross-Platform Aggregation  & \No   & \No   & \No   & \Ys   & \Ys   \\\hline
Training Acceleration       & \Ys   & \Ys   & \Ys   & \No   & \Ys   \\
MLOps                       & \Ys   & \Ys   & \No   & \No   & \Ys   \\
Open-Source Backend         & \No   & \No   & \Ys   & \Ys   & \Ys   \\
\end{tabular}
\caption{Functionality Comparison among On-Smartphone FL Systems.}
\label{tbl:fn-systems}
\end{table}

\subsubsection{Model Conversion}
To enable model creation in Python,
we begin by converting models into formats compatible with
Android (TensorFlow Lite or TFLite) and iOS (Core ML).
Core ML defines a fixed model structure and provides
the official converter CoreMLTools.
For TFLite, we standardized a model format, and then
developed a compliant TensorFlow converter.
This standardized format includes
four essential FL methods
(\textsf{train}, \textsf{infer}, \textsf{parameters},
and \textsf{restore}).

\subsubsection{Unified Training APIs}
\FedKit{} provides \textit{TFLite Trainer} and \textit{Core ML Trainer} to
train the converted models on Android and iOS devices,
utilizing GPU and NPU acceleration.
Moreover, both trainers expose unified APIs for
\textit{retrieving and setting model parameters,
    model fitting, and evaluation}.
On Android, these APIs invoke the TFLite interpreter to call
our \textit{standardized methods} defined in \textit{Model Conversion}.
However, on iOS, our experimentation revealed that
Core ML \textit{forbids} directly setting parameters, which could impede FL.
To overcome this constraint,
we apply a mostly undocumented method that
modifies the underlying ProtoBuf representations of Core ML models.
Specifically,
we employ Swift code generated from the relevant ProtoBuf definition files,
and navigate nested model definitions to access parameters on iOS.
Consequently, our unified training APIs exhibit comparable functionality on
both iOS and Android platforms.

\subsubsection{Cross-Platform Aggregation}
Aggregation necessitates
\textit{uniform parameter representations},
posing primary challenges in
retrieving and setting parameters for Core ML and TFLite.
1)~Core ML permits \textit{only} specific layers to be \textit{updatable} and
only provides their parameters post-training.
Thus, to obtain \textit{other layers'} parameters,
we implemented a solution using ProtoBuf manipulation.
This approach involves recording layer information
in \textit{Model Conversion} and
utilizing it during training.
2)~The TFLite interpreter only accepts inputs/outputs as maps from
\textit{names} to \textit{tensors}.
Therefore, during \textit{Model Conversion},
we assign index-based names to each parameter layer and
dynamically generate the concrete methods that accept these arguments.
During training, we call the methods with these index-based names to
access parameters.
Finally, these unified parameters enable seamless cross-platform aggregation.

\subsection{Flexible MLOps in Production}
\label{sec:mlops}
\begin{figure}
    \centering
    \includegraphics*[width=0.8\linewidth]{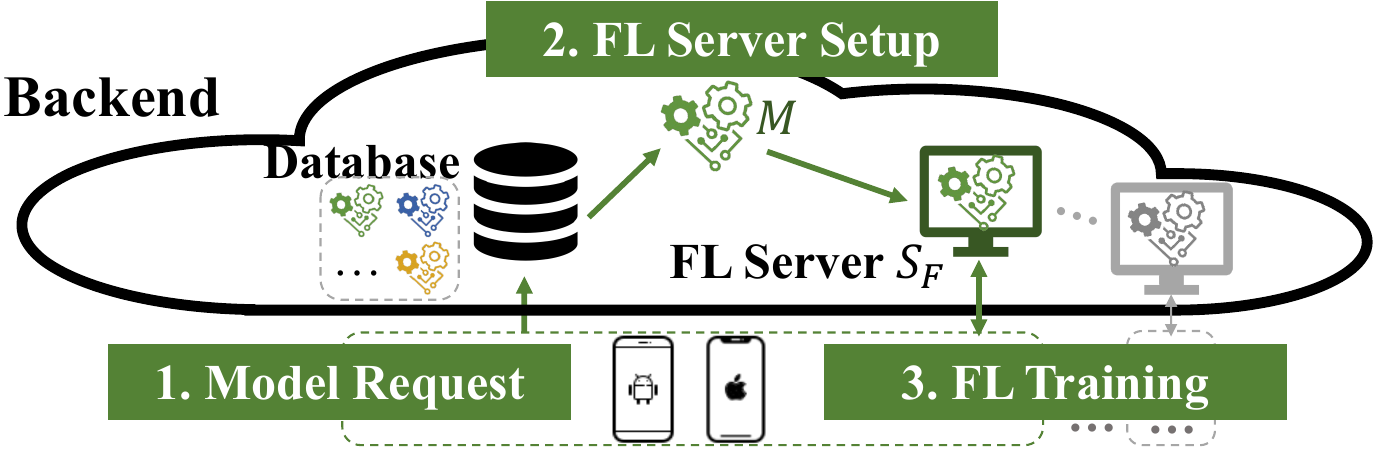}
    \vspace{-1mm}
    \caption{\FedKit{} FL Workflow.}
    \label{fig:fl-workflow}
    \vspace{-1mm}
\end{figure}

\newcommand{\model}{$M$}
\newcommand{\fs}{$S_\mathrm F$}
In production,
FL development faces challenges from
the lack of direct control over end devices.
\FedKit{} empowers researchers to deploy models and algorithms continuously (MLOps).
Leveraging our complete control of the self-hosted Backend,
\FedKit{}'s three-step FL workflow
facilitates continuous delivery and training,
as illustrated in Fig.~\ref{fig:fl-workflow}.

\subsubsection{Continuous Cross-Platform Model Delivery}
Traditionally, models for on-device training are \textit{embedded} into client apps.
However, this approach couples model delivery with app updates,
resulting in complexities when submitting apps to app stores and garnering user adoption.

Circumventing this complexity,
\FedKit{} enables \textit{continuous model delivery without app updates} by
decoupling models from clients through \textit{Model Request},
which allows new model deployment by uploading to the Backend.
Specifically, clients request the Backend for a model (TFLite/Core ML)
aligned with
their platform (Android/iOS) and training data type.
The Backend selects the appropriate model \model{} from the database and
responds with detailed model information.
Consequently, \textit{Model Request} delivers the TFLite or Core ML model
to clients for FL training.

\subsubsection{Customizable Continuous FL Training}
\FedKit{} manages continuous FL training by allowing \textit{multiple parallel FL training sessions}
through \textit{FL Server Setup}.
When clients request an FL Server to
train their chosen model \model{},
the Backend either reuses a suitable FL Server \fs{} if it exists,
or spawns a new one.
Each FL Server
operates as an independent Python subprocess of the Django Backend,
occupying its own port that
clients connect to for FL Training.
This dynamic approach ensures that
newly delivered models can be immediately trained with new FL Servers
without affecting ongoing ones.
Furthermore,
these FL Servers employ the Flower FL Framework~\cite{mathur2021ondevice} for
scheduling training and evaluation.
This decision empowers our FL Servers to leverage Flower's flexibility and
allow for FL algorithm customizations in Python.

\section{Live Demonstrations}

We demonstrate \FedKit{}'s effectiveness in two settings.\footnote{
    Demo video: \url{https://www.youtube.com/watch?v=TONTBkp_l6M}.
}

\subsubsection{Model Deployment on Demo Android/iOS App}
We demonstrate FL among
devices running a Flutter client app
and a laptop running a \FedKit{} Backend.
First, we demonstrate our seamless \textit{FL model pipeline}.
We convert a TensorFlow MNIST model,
and conduct \textit{normal} FL across an Android and an iOS device,
despite the heterogeneity.
Note that TFLite, Core ML, and the aggregation strategy determine the model
performance.
Second, to showcase \textit{MLOps},
we modify the model and deploy its new version.
As outlined in Table~\ref{tbl:demo-stats},
our telemetry shows that
the iOS device is over 5$\times$ faster in local training despite
having 0.5$\times$ RAM,
illustrating how \FedKit{} will provide real-world statistics to
enhance FL algorithm design.

\begin{table}
    \centering
    \setlength{\tabcolsep}{4pt}
\begin{tabular}{lllll}
Device      & System on a Chip  & Accel.        & RAM           & Time   \\\hline
Nova 9 Pro  & Snapdragon 778G   & OpenCL        & 8GB LPDDR5    & 3.583s \\
iPhone 13   & A15 Bionic, 4 GPU & CoreML        & 4GB LPDDR4X   & 0.656s \\
\end{tabular}
\caption{Configurations of Devices and Average Local Training Time Per Round
    (Two Local Epochs) in A Previous Demo Run.
}
\label{tbl:demo-stats}
    \vspace{-2mm}
\end{table}

\subsubsection{\FedCampus{}}
\begin{figure}
    \centering
    \includegraphics*[width=0.9\linewidth]{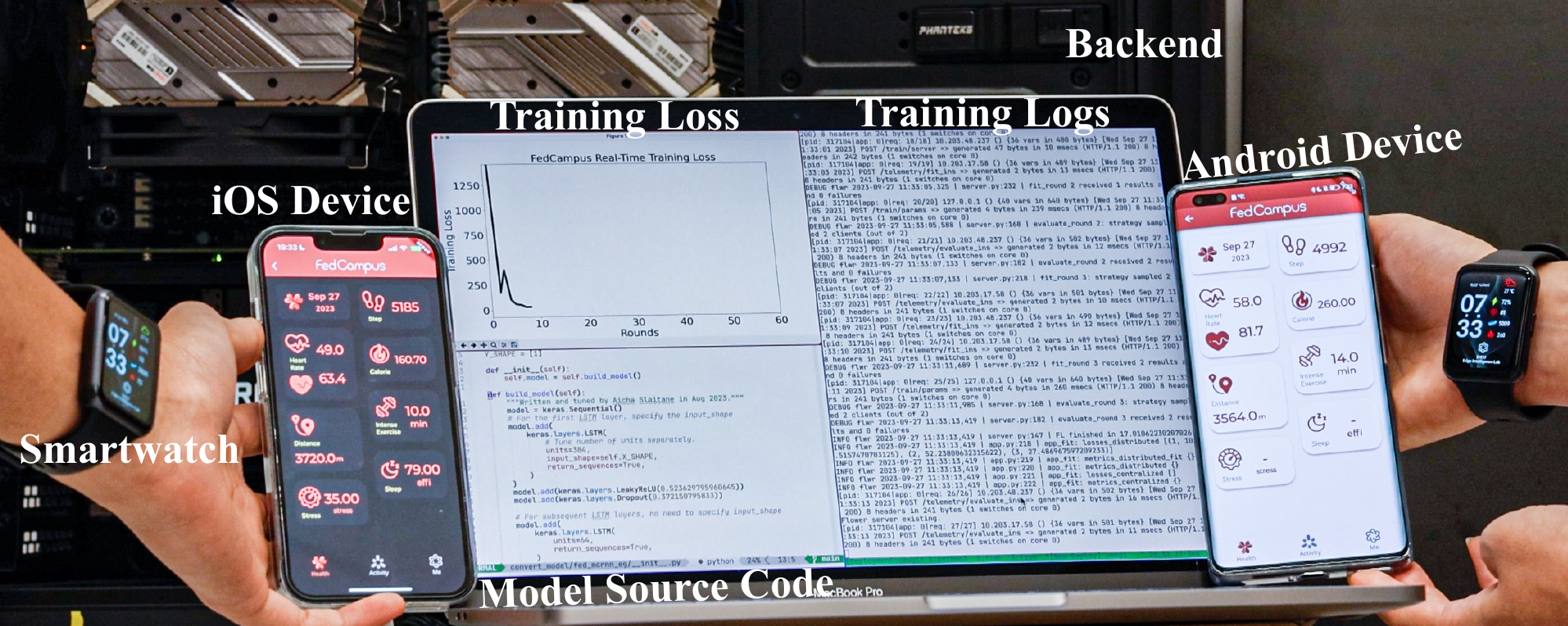}
       \vspace{-1mm}
    \caption{\FedCampus{} Experiment Setup on Duke Kunshan University Campus.}
    \label{fig:fedcampus}
        \vspace{-1mm}
\end{figure}

To harness real users' health data on university campuses,
we developed the \FedCampus{} Android/iOS app to leverage data from
participants' smartwatches to perform FL on
a sleep-duration prediction model.
As illustrated in Fig.~\ref{fig:fedcampus},
we showcase our self-hosted Backend,
and display the real-time logs and losses on a connected laptop.
The results of our cross-platform continuous training showcased
a significant reduction in the model's training loss,
demonstrating \FedKit{}'s effectiveness in real-world scenarios.

\bibliographystyle{IEEEtran}
\bibliography{main}

\end{document}